\documentclass[11pt]{IEEEtran}
\usepackage[table,xcdraw]{xcolor} 

\usepackage{pgf,tikz}
\usepackage{mathrsfs}
\usepackage[linesnumbered,ruled,vlined]{algorithm2e}
\usepackage{tabularx} 
\usepackage{caption} 
\usepackage{subcaption}
\usepackage{pgfplots}
\pgfplotsset{compat=1.14}
\usetikzlibrary{arrows}
\usepackage{multirow}
\usepackage{graphicx}
\usepackage[table,xcdraw]{xcolor}

\usepackage{subcaption}

\providecommand{\keywords}[1]{\textbf{\textit{Keywords---}} #1}

\mathchardef\mhyphen="2D 

\begin{document}

\title{AEGR: A simple approach to gradient reversal in autoencoders for network anomaly detection}
\author{Kasra Babaei, ZhiYuan Chen, Tomas Maul}
\date{}
\maketitle

\begin{abstract}
	Anomaly detection is referred to as a process in which the aim is to detect data points that follow a different pattern from the majority of data points. Anomaly detection methods suffer from several well-known challenges that hinder their performance such as high dimensionality. Autoencoders are unsupervised neural networks that have been used for the purpose of reducing dimensionality and also detecting network anomalies in large datasets. The performance of autoencoders debilitates when the training set contains noise and anomalies. In this paper, a new gradient-reversal method is proposed to overcome the influence of anomalies on the training phase for the purpose of detecting network anomalies. The method is different from other approaches as it does not require an anomaly-free training set and is based on reconstruction error. Once latent variables are extracted from the network, Local Outlier Factor is used to separate normal data points from anomalies. A simple pruning approach and data augmentation is also added to further improve performance. The experimental results show that the proposed model can outperform other well-know approaches. 
\end{abstract}

\keywords{network anomaly detection, high dimensionality, autoencoders (AEs), Local Outlier Factor (LOF), Gradient Reversal}

\section{Introduction}
\par In many real-world problems such as detecting fraudulent activities or detecting failure in aircraft engines, there is a pressing need to identify observations that have a striking dissimilarity compared to the majority. In medicine for instance, this discovery can lead to early detection of lung cancer or breast cancer. One area that is growing really fast is computer networks that plays a pivotal role in our daily life. Protecting networks from various threats such as network intruders is crucial. By using machine learning algorithms, it is possible to monitor and analyse the network and detect these threats almost instantly. However, when the number of observations that the method aims to detect is very small with respect to the whole data, these methods start to struggle, i.e., their performance debilitates. These observations are called anomalies (also known as outliers). The process whereby the aim is to detect data instances that deviate from the pattern of the majority of data instances is referred to as anomaly detection \cite{nicolau2018learning}. Depending on the application domain, anomalies can occur due to various causes such as human error, system failure, or fraudulent activities. 

\par Traditional anomaly detection methods, e.g., density-based or distance-based, are found to be less effective and efficient when dealing with large-scale, high-dimensional and time-series datasets \cite{Munir20191991}\cite{ma2013parallel}\cite{zhang2018survey}. Moreover, most  of these approaches in the literature often require large storage space for the intermediate generated data \cite{sun2018learning}. 

\par Moreover, parameter tuning for classical approaches such as clustering models in large-scale datasets is a difficult task \cite{8620681}. Consequently, the dimensionality of such datasets should be reduced before applying anomaly detection methods. This is achieved by performing a pre-processing step known as dimensionality reduction, which tries to remove irrelevant data while keeping important variables and transforming the dataset into a lower dimension. Besides high dimensionality, lack of labelled datasets is another  problem in this context. Many supervised learning algorithms have been employed to detect anomalies; however, an essential requirement of a supervised learning approach is labelled data. While availability of labelled datasets is often a problem, labelling can also be costly, time-consuming, and requires a field expert \cite{Hodge2004}. Lastly, in many application domains, such as telecommunication fraud and computer network intrusion, companies are often very conservative and protective of their data due to privacy issues and tend to resist providing their data \cite{sun2018learning}. 

\par In this paper, the aim is to tackle the mentioned problems by employing an unsupervised approach in which high-dimensional data is compressed into lower dimensionality, and a density-based method is then applied to the transformed data in order to detect network anomalies. In particular, a deep autoencoder (AE) network is used to create low-dimensional representations, and next, Local Outlier Factor (LOF) \cite{Breunig:2000:LID:335191.335388} is utilised for separating normal data instances from anomalies. Although autoencoders have shown promising results, their performance weakens when the dataset contains noise and anomalies \cite{Qi20146716}. To overcome this issue, various approaches have been proposed such as denoising autoencoders or using a loss function which is insensitive to anomalies and noise \cite{zhou2017anomaly}. 

\par The main contributions of this paper, as will be elaborated in subsequent sections, are the following:
\begin{itemize}
	\item First, unlike other similar approaches that require a noise and anomaly-free training set, the proposed model in this paper is insensitive to anomalies; therefore, our approach is needless of anomaly-free training sets.
	\item Second, the proposed method is robust to large datasets, and is particularly good with high-dimensional network datasets.
	\item Third, the method is capable of working with unlabelled datasets. 
\end{itemize}

\par The proposed model is tested on 8 different datasets including 5 well-known network datasets. Evaluation of experimental results show that the proposed model can improve the performance significantly and is superior to the stand-alone LOF and two other state-of-the-art approaches. The rest of this paper is organised as follows. Section \ref{sec:related_studies} reviews previous works, while section \ref{sec:proposed_method} explains the proposed model. Experimental results are presented and discussed in section \ref{sec:experiment}, and subsequently the paper concludes with section \ref{sec:conclusion}.

\section{Related Studies}
\label{sec:related_studies}
\par There are various methods for anomaly detection and several studies have categorised them into different groups. For instance, in \cite{Tang2017171}, authors categorised them into the following four groups: distribution-based, distance-based, clustering-based, and density-based methods. Arguably, the most widely accepted categorisation is based on the type of supervision used, i.e., unsupervised, semi-supervised and supervised \cite{Munir20191991}. Except for unsupervised methods, labelled data is required for training models that are semi-supervised or supervised, and as explained earlier, labelling data brings various challenges; therefore, unsupervised models are more favourable \cite{Hodge2004}. 

\par Traditional anomaly detection methods try to search the entire dataset and detect anomalies which will result in discovering global anomalies. However, in many real-world problems the data is incomplete and often the application requires a local neighbourhood search for identifying anomalies \cite{su2017n2dlof}. One of the approaches that aims at measuring the outlierness of each data instance based on the density of its local neighbourhood is LOF. However, traditional methods such as LOF fail to achieve an acceptable result when applied to large and high-dimensional datasets, and moreover they tend to require large storage capacity in this context \cite{zhang2018survey}\cite{ma2013parallel}.
 
\par To overcome the aforementioned problems, several approaches have been proposed in which high-dimensional data is transformed into a low-dimensional space while trying to avoid loss of crucial information. Next, anomaly detection is carried out on the low-dimensional data. 

\par Recently, autoencoders have been widely employed for the purpose of reducing dimensionality of large datasets. An autoencoder refers to an unsupervised neural network that tries to reconstruct its input at the output layer \cite{hinton2006reducing}. It is made of two main parts, namely an encoder and a decoder. The encoder tries to convert the input features into a different space with lower dimension while the decoder attempts to reconstruct the original feature space using the output of the decoder, which is known as the bottleneck or discriminative layer. In terms of network structure, autoencoders come in various types such as under-complete, over-complete, shallow or deep. A comprehensive review can be found in \cite{charte2018practical}. Because of its ability to encode data without losing information it has been widely employed in the literature. Unlike other dimensionality reduction methods such as Principal Component Analysis (PCA) that use linear combinations, AEs generally perform nonlinear dimensionality reduction and, according to the literature, perform better than PCAs \cite{wang2016auto}. Authors in \cite{sakurada2014anomaly} used AE for anomaly detection and compared its performance to linear PCA and kernel PCA on both synthetic and real data, and based on the result, they concluded that AE can extract more subtle anomalies than PCA. Another disadvantage of statistical algorithms such as PCA or Zero Component Analysis (ZCA) is that as the dimensionality increases, more memory is required to calculate the covariance matrix \cite{yousefi2017autoencoder}.

\par A well-trained autoencoder should generate small reconstruction errors for each data point; however, autoencoders fail at replicating anomalies because their patterns deviate from the pattern that the majority of data instances follow. In other words, the Reconstruction Error (RE) of anomalies is greatly above the RE of normal data. In some studies such as \cite{aygun2017network}, the authors used a threshold-based classification using the reconstruction error as a score to separate normal data from anomalies, i.e., the data instances that have a reconstruction error above the threshold are identified as anomalies while anything below the threshold is considered as normal. In another similar approach, the authors in \cite{schreyer2017detection} generated an anomaly score based on reconstruction error and individual attribute probabilities in large-scale accounting data. 

\par The authors in \cite{chen2017outlier} proposed an approach in which ensembles of AEs were used to improve the robustness of the network. In their approach, named Randomised Neural Network for anomaly Detection (RandNet), instead of using fully connected layers, the connections between layers are randomly altered, and anomalies are separated from normal data using the reconstruction error. Their approach showed superior performance compared to four traditional anomaly detection methods including LOF.

\par To enhance the performance of AEs, authors in \cite{sun2018learning} added new regularisers to the loss function that encourage the normal data to create a dense cluster at the bottleneck layer while the anomalies tend to stay outside of the cluster. Next, they employed various Once Class Classification (OCC) methods such as LOF, Kernel Density Estimation (KDE), and One Class SVM (OCSVM) to divide the data into anomalies and normal data. They also investigated the influence of altering the training set size on the performance of their models and the result showed consistency across different training sizes. 

\par As the number of hidden layers increases, the backpropagated gradients to the lower layers tend to attenuate, which is known as the vanishing gradient problem, and culminating in the weights of lower layers showing limited change. To overcome the vanishing gradient issue and discover better features, a pretraining phase was proposed in \cite{yousefi2017autoencoder} in which a stacked Restricted Boltzmann Machine (RBM) was used to obtain suitable weights for initialising the AE. The authors applied their model to the NSL-KDD dataset and achieved high accuracy ($83.34\%$).

\par The performance of AEs diminishes when datasets contain anomalies and/or noise which is very prevalent in real-world datasets. By using Denoising AEs (DAEs), it is possible to enhance the accuracy of the network. DAE is an extension of AE in which the network is trained on an anomaly and noise-free set while random noise is added to the input and the AE tries to regenerate the original input, i.e., without the added noise. Authors in \cite{sakurada2014anomaly} used a DAE to obtain meaningful features and decrease the dimensionality of data. Their result proved that DAE is superior to statistical methods such as PCA. 

\par DAEs require an anomaly and noise-free training set; however, such a training set is not always available. In related work, \cite{zhou2017anomaly}, the authors proposed an AE, called Robust Deep Autoencoder (RDA), that can extract satisfactory features without having access to an anomaly or noise-free training set. Inspired by Robust PCA (RPCA), they added a penalty-based filter layer to the network that uses either $L_1$ or $L_{2,1}$ norms, and managed to separate anomalies and noise from normal data. 

\par A common criterion used in AEs is Mean Square Error (MSE). As mentioned before, anomalies and noise tend to cause greater reconstruction error, i.e., larger MSE; consequently, the network carries out a substantial weight update. Therefore, it debilitates the accuracy of AEs as they tend to learn to regenerate anomalies and noise. A possible remedy to this problem is using a criterion which is insensitive to anomalies and noise. Authors in \cite{Qi20146716} proposed a new approach, called Robust Stacked AutoEncoder (R-SAE), in which they used the maximum correntropy criterion to prevent substantial weight updates and make the model robust to anomalies and noise. They tested their model on the MNIST dataset contaminated with non-Gaussian noise, and achieved $39\%$-$63\%$ lower REs compared to what was obtained from a Standard Stacked AutoEncoder (S-SAE).

\section{Proposed Method}
\label{sec:proposed_method}
\par In the previous section, different approaches for extracting meaningful features from high-dimensional datasets were reviewed and also previous anomaly detection models were explained. In this section, we will explain how our model robustly transforms high-dimensional data into a low-dimensional space and detects anomalies.

\subsection{Local Outlier Factor}
\label{subsec:local_outlier_factor}
\par LOF is a density-based anomaly detection method in which the assumption is that anomalies tend to stay outside of dense neighbourhoods because of their peculiar characteristics that distance them from inliers \cite{Breuniq200093}. The method generates a score that shows the outlierness of each data point based on its local density. A low score indicates that the query point is an inlier while a high score shows that the query point is an anomaly. The algorithm has one parameter, $MinPts$, which is the minimum number of neighbours that each data point is expected to have inside its neighbourhood. 

\par It is possible to divide the anomaly detection process of LOF into three steps. In the first step, LOF tries to find the minimum distance, called $k{\mhyphen}distance$, to hold at least $MinPts$ neighbours. Next, the algorithm measures the reachability distance defined as:
\begin{equation}
	reach{\mhyphen}dist_{k}(p,o)=max\{k{\mhyphen}distance(o),d(p,o)\}
	\label{equ:reach-dist}
\end{equation}
which is equal to $k{\mhyphen}distance$ of point $o$ if the query point is also inside point $o$'s neighbourhood, otherwise it is equal to the actual distance between the two points. In the second step, the local reachability distance (LRD), which is the inverse of the average reachability distance of the data point $p$ from its neighbours, is measured. LRD is defined as:
\begin{equation}
	{\scriptstyle LRD_{MinPts}(p)=1/}\Bigg(\frac{\mathlarger{\sum}_{\scriptscriptstyle{o\in{N_{MinPts}(p)}}}{\scriptscriptstyle{reach{\mhyphen}dist_{MinPts}(p, o)}}}{\scriptscriptstyle{|N_{MinPts}(p)|}}\Bigg)
	\label{equ:lrd}
\end{equation}
in which $MinPts$ denotes the number neighbours in data point $p$'s neighbourhood that is used to detect anomalies. In the final step, the LRD of the query point is compared with the LRD of its neighbours using the following equation:
\begin{equation}
	LOF_{MinPts}(p)=\frac{{\mathlarger{\sum}_{o\in{N_{MinPts}(p)}}{\frac{lrd_{MinPts}(o)}{lrd_{MinPts}(p)}}}}{|N_{MinPts}(p)|}
	\label{equ:LOF}
\end{equation}

If the density of point $p$ is very close to its neighbours, then the value of LOF stays around $1$ while for inliers it is less than $1$ and for anomalies it is greater than $1$. It is worth mentioning that it is very common to apply a simple threshold-based clustering here to separate anomalies.

\subsection{AutoEncoders}
\label{subsec:autoencoders}
\par
\begin{figure}[!h]
\begin{center}
	
	\begin{tikzpicture}[scale=0.8,line cap=round,line join=round,>=triangle 45,x=1.0cm,y=1.0cm]
	\clip(-4.83199028842192,-3.350560119718393) rectangle (5.188918188289323,2.667106384609909);
	\draw [line width=1.2pt] (-4.,-2.) circle (0.32802438933713446cm);
	\draw [line width=1.2pt] (-4.,2.) circle (0.3124099870362663cm);
	\draw [line width=1.2pt] (-4.,1.) circle (0.32802438933713435cm);
	\draw [line width=1.2pt] (-4.,0.) circle (0.3124099870362663cm);
	\draw [line width=1.2pt] (-4.,-1.) circle (0.3124099870362663cm);
	\draw [line width=1.2pt] (-2.,1.) circle (0.3280243893371345cm);
	\draw [line width=1.2pt] (-2.,0.) circle (0.3124099870362662cm);
	\draw [line width=1.2pt] (-2.,-1.) circle (0.31240998703626616cm);
	\draw [line width=1.2pt] (0.,0.) circle (0.3124099870362662cm);
	\draw [line width=1.2pt] (2.,1.) circle (0.32802438933713435cm);
	\draw [line width=1.2pt] (2.,0.) circle (0.3124099870362663cm);
	\draw [line width=1.2pt] (2.,-1.) circle (0.3124099870362663cm);
	\draw [line width=1.2pt] (4.,-2.) circle (0.32802438933713446cm);
	\draw [line width=1.2pt] (4.,2.) circle (0.3124099870362663cm);
	\draw [line width=1.2pt] (4.,1.) circle (0.32802438933713435cm);
	\draw [line width=1.2pt] (4.,0.) circle (0.3124099870362663cm);
	\draw [line width=1.2pt] (4.,-1.) circle (0.3124099870362663cm);
	\draw [->,line width=0.8pt] (-3.671975610662866,1.) -- (-2.3272978670955706,0.9781801421936286);
	\draw [->,line width=0.8pt] (-3.6875900129637342,0.) -- (-2.312409987036266,0.);
	\draw [->,line width=0.8pt] (-3.687590012963734,-1.) -- (-2.311616059483331,-1.022258289963095);
	\draw [->,line width=0.8pt] (-1.6875900129637338,0.) -- (-0.3124099870362662,0.);
	\draw [->,line width=0.8pt] (0.3124099870362662,0.) -- (1.6727021329044296,0.9781801421936286);
	\draw [->,line width=0.8pt] (0.3124099870362662,0.) -- (1.6875900129637338,0.);
	\draw [->,line width=0.8pt] (0.3124099870362662,0.) -- (1.6903305043977526,-1.0412892660802997);
	\draw [->,line width=0.8pt] (2.3257776205028176,0.9616732211173156) -- (3.67270213290443,0.9781801421936286);
	\draw [->,line width=0.8pt] (2.312409987036266,0.) -- (3.6875900129637342,0.);
	\draw [->,line width=0.8pt] (2.312409987036266,0.) -- (3.672809219099011,-2.023370770064356);
	\draw [line width=0.8pt] (2.3118015931053515,-1.0194875995690844)-- (3.687590012963734,2.);
	\draw [->,line width=0.8pt] (2.3118015931053515,-1.0194875995690844) -- (3.687590012963734,-1.);
	\draw [line width=2.pt] (-4.580833035493351,-2.6108687372438015)-- (-0.9977224069976642,-2.6004223505718023);
	\draw [line width=2.pt] (0.976751104527751,-2.591669597353602)-- (4.559861733023438,-2.581223210681603);
	\draw [line width=2.pt] (-0.49585612643783666,-2.616490163162033)-- (0.5028204423655828,-2.616490163162033);
	\draw [line width=0.8pt] (-3.687590012963734,2.)-- (-2.3272978670955706,0.9781801421936286);
	\draw [line width=0.8pt] (-3.6875900129637342,0.)-- (-2.3272978670955706,0.9781801421936286);
	\draw [line width=0.8pt] (-3.687590012963734,-1.)-- (-2.3272978670955706,0.9781801421936286);
	\draw [line width=0.8pt] (-3.672614412416798,-2.02046159922395)-- (-2.3272978670955706,0.9781801421936286);
	\draw [line width=0.8pt] (-3.687590012963734,2.)-- (-2.312409987036266,0.);
	\draw [line width=0.8pt] (-3.671975610662866,1.)-- (-2.312409987036266,0.);
	\draw [line width=0.8pt] (-3.687590012963734,-1.)-- (-2.312409987036266,0.);
	\draw [line width=0.8pt] (-3.672614412416798,-2.02046159922395)-- (-2.312409987036266,0.);
	\draw [line width=0.8pt] (-3.687590012963734,2.)-- (-2.311616059483331,-1.022258289963095);
	\draw [line width=0.8pt] (-3.671975610662866,1.)-- (-2.311616059483331,-1.022258289963095);
	\draw [line width=0.8pt] (-3.672614412416798,-2.02046159922395)-- (-2.311616059483331,-1.022258289963095);
	\draw [line width=0.8pt] (-1.672480652362455,0.981804480686803)-- (-0.3124099870362662,0.);
	\draw [line width=0.8pt] (-1.688129116017763,-1.018345346116602)-- (-0.3124099870362662,0.);
	\draw [line width=0.8pt] (2.312409987036266,0.)-- (3.67270213290443,0.9781801421936286);
	\draw [line width=0.8pt] (2.3118015931053515,-1.0194875995690844)-- (3.67270213290443,0.9781801421936286);
	\draw [line width=0.8pt] (2.3257776205028176,0.9616732211173156)-- (3.6875900129637342,0.);
	\draw [line width=0.8pt] (2.3118015931053515,-1.0194875995690844)-- (3.6875900129637342,0.);
	\draw [line width=0.8pt] (2.3257776205028176,0.9616732211173156)-- (3.687590012963734,-1.);
	\draw [line width=0.8pt] (2.312409987036266,0.)-- (3.687590012963734,-1.);
	\draw [line width=0.8pt] (2.3257776205028176,0.9616732211173156)-- (3.672809219099011,-2.023370770064356);
	\draw [line width=0.8pt] (2.3118015931053515,-1.0194875995690844)-- (3.672809219099011,-2.023370770064356);
	\draw [line width=0.8pt] (2.3257776205028176,0.9616732211173156)-- (3.687590012963734,2.);
	\draw [->,line width=0.8pt] (2.319832405866583,-0.011039630830771835) -- (3.6847041472632425,1.993666296669693);
	\begin{scriptsize}
	\draw[color=black] (-2.7,-2.8) node {$Encoder$};
	\draw[color=black] (2.8,-2.8) node {$Decoder$};
	\draw[color=black] (0.0,-2.8) node {$Bottleneck$};
	\end{scriptsize}
	\end{tikzpicture}
\end{center}
\caption{The structure of a deep undercomplete autoencoder}
\label{fig:structure_of_ae}
\end{figure}
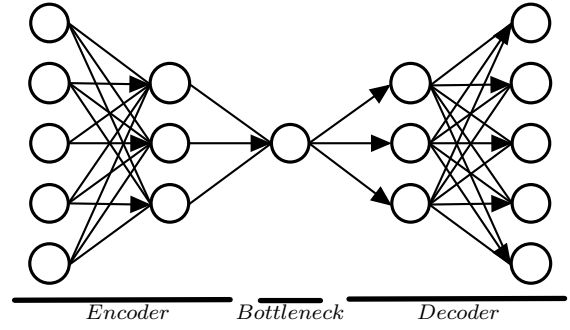

 Autoencoders are unsupervised artificial neural networks that contain two components \cite{hinton2006reducing}. The first component, known as the \textit{encoder}, tries to transform the high-dimensional input data into a low-dimensional feature space, known as the \textit{bottleneck}, while the second component, known as the \textit{decoder}, attempts to reconstruct the input data from the bottleneck. The difference between the reconstructed and input data is called the \textit{reconstruction error}. In each training iteration, the network measures the reconstruction error, computes the gradient of the error with respect to network parameters (e.g. weights), and  backpropagates these gradients through the network in order to update the weights to minimise the reconstruction error, i.e., to increase the resemblance between the generated output and the input. AEs can adopt various structures. In Figure \ref{fig:structure_of_ae}, the structure of an AE, known as Stacked Autoencoder (SAE), is shown in which multiple layers are stacked to form a deep neural network.  

\par
\begin{figure}[!h]
	\begin{center}
		\begin{tikzpicture}[line cap=round,line join=round,>=triangle 45,x=1.0cm,y=1.0cm]
		\clip(-3,-0.6) rectangle (3,.6);
		\draw [line width=0.8pt] (-2.,0.) circle (0.5);
		\draw [line width=0.8pt] (0.,0.) circle (0.5);
		\draw [line width=0.8pt] (2.,0.) circle (0.5);
		\draw [->,line width=1.pt] (-1.5,0.) -- (-0.5,0.);
		\draw [->,line width=1.pt] (0.5,0.) -- (1.5,0.);
		\begin{scriptsize}
		\draw[color=black] (-2,0) node {$x$};
		\draw[color=black] (0,0) node {$z$};
		\draw[color=black] (2,0) node {$y$};
		
		\draw[color=black] (-1.2,0.2) node {$u$};
		\draw[color=black] (0.85,0.2) node {$v$};
		\end{scriptsize}
		\end{tikzpicture}
	\end{center}
	\caption{The structure of a basic autoencoder}
	\label{fig:basic_ae}
\end{figure}
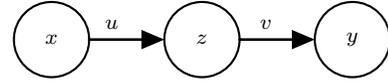

 As shown in Figure \ref{fig:basic_ae}, the most basic AE with only one hidden layer tries to transform input $x$ into latent vector $z$ using an encoder represented by function $u$. Next, the latent vector $u$ is reconstructed by a decoder represented as function $v$ into output $y$ where the dissimilarity between $y$ and $x$ is called reconstruction error. Having a training set, $D_u=\{x_1,x_2, x_3,...,x_n\}$, where $n$ is the number of data points in $D$ and $x_i$ is the $i$th data point with $m$ features, the encoder is then defined as:

\begin{equation}
	z = u(x) = s(W x + b)
	\label{equ:encoder}
\end{equation}

while the decoder is defined as:

\begin{equation}
	y = v(z) = s'(W' z + b')
	\label{equ:decoder}
\end{equation}

where both $s$ and $s'$ represent the activation functions that are most often non-linear, $W$ and $W'$ denote the weight matrices while $b$ and $b'$ represent the bias vectors. 

\par It is worth mentioning that, in general, certain restrictions or regularisation techniques should be applied to an AE in order to prevent the network from learning the identity function. Otherwise, the network will simply copy the input through the network. A common solution to overcome this issue is by using a denoising AE. In a DAE, the network tries to reconstruct inputs that are partially corrupted. Bottlenecks and sparsity constraints applied to the main hidden layer are additional ways to avoid trivially learning the identity function.

\subsection{AEGR: The Proposed Model for anomaly Detection in Large Datasets} 
\par The proposed AEGR (autoencoder with gradient reversal) approach consists of two main components. The first component is an AE which transforms the high dimensional data into compressed data while preserving important latent variables for the following component. The second component employs a basic LOF approach on the features obtained from the first component to separate anomalies from normal data points. 
\begin{algorithm}[t]

\SetAlgoLined
 
 \SetKwInOut{Input}{input}\SetKwInOut{Output}{output}
 \Input{$D$: a dataset with $n\times m$ dimension \\
 $e \leftarrow$ number of epochs}
 \Output{$D'$: a dataset with $n\times m'$ dimension} 
 \BlankLine
 
 initialise: $k \leftarrow$ GR start point\;
 $LGS =$ list of gradient scores\;
 \For{each epoch}{
	 \For{each point/mini-batch in $epoch_j$}{
		 \If{$j > k$}{
		 	\For{each layer}{
			 	\If{bottleneck}{
			 		$GS_j \leftarrow ||n||_F$\\
			 		$LGS \leftarrow GS_j$
			 	}
			}
		}
	 }
	 $Max_{GS} \leftarrow$ Max $GS$ in $LGS$ ;\\
	Bacpropagate inverted $Max_{GS}$;\\
	Shuffle the batches;
 }
 $D' \leftarrow$ the bottleneck of the last epoch
\caption{AEGR}
\label{alg:AEGD}
\end{algorithm}
\par According to the literature \cite{Qi20146716}, the performance of AEs diminishes when the data exhibits noise and anomalies. To explain further, the network tries to reduce the reconstruction error by learning how to reconstruct noise and anomalies. During the training phase, anomalies tend to produce a greater reconstruction error as their patterns deviate from the distribution that the majority of data points follow. The network thus disproportionately backpropagates the corresponding error gradients and performs a substantial weight update to be able to reproduce these patterns with lower error. However, in the context of anomaly detection, this is not desired. In fact, in previous works, various approaches have tried to prevent the network from becoming more accurate in reproducing anomalies. Most of these approaches such as DEA require access to a noise and anomaly-free set for training the network while our model is needless of such a training set. Authors in \cite{ganin2014unsupervised} tried to add a reversal layer to their network for the purpose of domain adaption. The approach proposed in this work is different in several ways including the context within which it is applied, and the fact that unlike the approach in \cite{ganin2014unsupervised}, it is based on the reconstruction error.



\par In AE with gradient reversal (AEGR), the AE component tries to make the network insensitive to anomalies by manipulating gradients. First, as shown in Algorithm \ref{alg:AEGD}, at each epoch a gradient score (GS) is given to each data point (or each mini-batch in the case of using mini-batch gradient descent) based on the gradients in the bottleneck. This is measured by using the Frobenius norm which is defined as:
\begin{equation}
	||X||_F = \sqrt{\sum\limits_{i=1}^k\sum\limits_{j=1}^l|x_{i,j}|^2}
	\label{equ:l2}
\end{equation}
where $X$ denotes a $k\times j$ matrix and $x_{ij}$ represents the element at the $i$th row and $j$th column. At the bottleneck layer, the Frobenius norm of each node is computed. The bottleneck is the layer that latent variables are extracted from to be used in LOF; therefore, the network should be penalised based on the magnitude of its gradients in this layer rather than every layer. Then, the norm of layer $l$ is measured and stored as:
\begin{equation}
	\{l_1, l_2, l_3, ..., l_n\}
	\label{equ:gradient_set}
\end{equation}
where $n$ denotes the number of data points (or mini-batches in case of using mini-batch gradient descent) in the network. The approach has a single parameter, $k$, which is the epoch number when the network starts reversing its gradients. 

\par Assuming that anomalies produce greater REs, at each epoch, the gradient of the data point that holds the largest $GS$ is picked, and the inverse of its gradient is used to perform a new weight update, i.e., reverting the substantial weight update caused by this data point. In order to avoid reverting the gradients of the same data points at each epoch, batches are shuffled before carrying out the next epoch.

\par In the second stage in which LOF is used to separate anomalies from normal data points, three different approaches are proposed. The first approach merely uses the latent variables that are extracted from the previous stage without any alternation while in the second approach, a threshold-based clustering method based on the reconstruction error of the training set is applied to prune data points that have generated an error above the mean value. The assumption behind this approach is that the training set becomes more homogeneous and consequently LOF can perform more robustly. However, this approach also reduces the number of training points that can weaken the performance of one-class classifiers \cite{Khan2010}. Therefore, a third approach is proposed in which the pruned training data is augmented by adding random Gaussian noise. The assumption is that data augmentation can homogeneously increase the size of the training set and improve anomaly detection performance.

\section{Experiments}
\label{sec:experiment}
\par In this section, the performance of our model is presented, evaluated and compared with 1) the basic stand-alone LOF algorithm; 2) a deep AE whose reconstruction error is used for separating anomalies; and 3) a deep AE whose latent variables are fed to LOF for detecting anomalies. One of the widely used evaluation metrics in this area is Receiver Operating Characteristics (ROC) AUC; however, authors in \cite{Provost:1997} claim that when the data set is highly imbalanced, ROC AUC is not a suitable metric and a better alternative would be the area under the Precision-Recall curve (PR) AUC, in particular, when working with high dimensional data in which the positive class, i.e, anomalies, is more important than the negative class, i.e., normal points. Nonetheless, no single evaluation metric dominates others; therefore, both ROC AUC and PR AUC were used for evaluation.

\subsection{Datasets}
\label{subsec:datasets}
\par As presented in Table \ref{tab:datasets_details}, 8 datasets were used that are publicly available and widely used in this context. It is worth mentioning that 5 network datasets that are very common in the domain of network anomaly detection were used besides 3 non-network datasets. Testing the model on various datasets makes it possible to evaluate performance more thoroughly. It is worth noting that the categorical features of two datasets, namely NSL-KDD and UNSW-NB15, were preprocessed by applying a one-hot-encoder. Four datasets, namely Spambase, InternetAds, Arrhythmia and CTU13, have no separate training and test sets; therefore, $60\%$ of each dataset was used for training while the rest was evenly split for validation and testing. For the remaining five datasets that come with a separated training and test set, $20\%$ of the training set was used for validation. Also, as suggested in \cite{Cao20193074}, the training sets of UNSW-NB15 and NSL-KDD are substantially larger than other datasets; therefore, only $10\%$ of the training set was used. In this paper, it was found necessary to apply the same size reduction approach to the CTU13 and Shuttle dataset. The details of each dataset are shown in Table \ref{tab:datasets_details}.

\begin{table}
\caption{Details of the 8 datasets used in the experiments}
\label{tab:datasets_details}
\resizebox{\columnwidth}{!}{%
\begin{tabular}{l|c|c|c|c}
Data set name & \begin{tabular}[c]{@{}c@{}}No. of \\ features\end{tabular} & \begin{tabular}[c]{@{}c@{}}Training \\ set\end{tabular} & \begin{tabular}[c]{@{}c@{}}Validation \\ set\end{tabular} & \begin{tabular}[c]{@{}c@{}}Testing \\ set\end{tabular} \\ \hline
PenDigits & 16 & 1247 & 312 & 727 \\
Shuttle & 9 & 3267 & 817 & 14500 \\
Arrhythmia & 259 & 251 & 83 & 86 \\
\hline
Spambase & 57 & 2759 & 919 & 923 \\
InternetAds & 1558 & 1414 & 471 & 474 \\
NSL-KDD(Probe) & 122 & 6319 & 1580 & 12132 \\
NSL-KDD(DoS) & 122 & 9060 & 2266 & 17169 \\
NSL-KDD(R2L) & 122 & 6183 & 1546 & 12598 \\
NSL-KDD(U2R) & 122 & 5428 & 1358 & 9778 \\
UNSW-NB15(Fuzzers) & 196 & 5934 & 1484 & 74184 \\
UNSW-NB15(Analysis) & 196 & 4640 & 1160 & 58000 \\
UNSW-NB15(Backdoor) & 196 & 4619 & 1155 & 57746 \\
UNSW-NB15(DoS) & 196 & 5460 & 1366 & 68264 \\
UNSW-NB15(Exploits) & 196 & 7151 & 1788 & 89393 \\
UNSW-NB15(Generic) & 196 & 7680 & 1920 & 96000 \\
UNSW-NB15(Reconnaissance) & 196 & 5319 & 1330 & 66491 \\
UNSW-NB15(Shellcode) & 196 & 4570 & 1143 & 57133 \\
UNSW-NB15(Worm) & 196 & 4584 & 1146 & 56130 \\
CTU13-13(Virut) & 40 & 4315 & 1438 & 1440 \\
CTU13-10(Rbot) & 38 & 7331 & 2443 & 2445 \\
CTU13-09(Neris) & 41 & 12895 & 4298 & 4301 \\
CTU13-08(Murlo) & 40 & 8045 & 2681 & 2683 \\
\hline
\end{tabular}%
}
\end{table}

\par The following datasets were obtained from the UCI Machine Learning Repository: PenDigits, Shuttle, Spambase, and InternetAds \cite{UCI_repo}. The CTU13-08 dataset was released in 2011, which is a botnet dataset and publicly available \cite{GARCIA2014100}. The NSL-KDD dataset is a new version of its predecessor, KDD'99, in which some of the intrinsic issues of its old version, mentioned in \cite{tavallaee2009}, are resolved. This dataset has 41 features in which 3 of them are categorical and after applying a one-hot-encoder, the number of features increased to 122. A similar preprocessing step was carried out on the UNSW-NB15 dataset \cite{UNSW-NB15} with 3 categorical features, which increased the number of features from 42 to 190. While 5 datasets contain only a single type of anomaly, three network datasets, namely UNSW-NB15, NSL-KDD and CTU13, include different types of anomalies (i.e., network attacks). The experiment was carried out on each type of these attacks.

\subsection{AEGR Architecture and Parameters}
\label{subsec:ae_architecture}

\par The AE architecture varies slightly based on the dataset it is applied on; however, the same architecture was used for both AE and AEGR. The number of layers was set to 5 for all the datasets as suggested in \cite{ERFANI2016121} with the number of nodes in the bottleneck being set to $m=[1+\sqrt{n}]$, where $n$ is the number of features of the input \cite{CaoVan2016}. Table \ref{tab:aegr_architecture} shows the number of nodes in the bottleneck for each dataset. 

\par For datasets with more than $2000$ instances, the mini-batch size was set to $64$ and $16$ otherwise. To avoid overfitting, a simple early stopping heuristic was implemented to stop the training process when the network was no longer learning after a certain number of iterations or when the learning improvement was insignificant. 

\par The loss function used in this experiment was $Smoothl1Loss$, which is essentially a combination of $L2$ and $L1$ terms, i.e., it uses $L2$ if the absolute element-wise error is less than 1 and $L1$ term if not. To minimise the loss function, Stochastic Gradient Descent (SGD) was used.

\par All the datasets where normalised into the range $[-1,+1]$, and the hyperbolic tangent activation function was used for all the layers except the last layer. The activation function is defined as:
\begin{equation}
	f_{tanh}(x) = \frac{e^x - e^{-x}} {e^x + e^{-x}}
	\label{equ:tangent}
\end{equation} 
where the output range is $(-1,+1)$.

\begin{table}
	\caption{The size of the middle layer for each dataset}
	\label{tab:aegr_architecture}
		\footnotesize
		\begin{tabularx}{\linewidth}{l l}
			\hline
			Dataset's name& \begin{tabular}[c]{@{}l@{}}Number of nodes in\\ the bottleneck\end{tabular}\\
			\hline
			PenDigits & 5 \\
			Shuttle & 4 \\
			Spambase & 8 \\
			InternetAds & 40 \\
			Arrhythmia & 17 \\
			NSLKDD & 12 \\
			UNSW-NB15 & 15 \\
			CTU13 & 7 \\
			\hline
		\end{tabularx}
\end{table} 

\subsection{LOF Parameters}
\label{subsec:lof_parameters}
\par The LOF algorithm has only one parameter, i.e., $k$, that needs to be set. This value indicates what is the minimum number of neighbours that each data instance requires to have when computing its density. Throughout our experiment this $k$ was set to a constant value. 

\subsection{Analysis and Discussion}
\label{subsec:discussion}
\begin{table*}
\centering
\caption{The PR AUC and ROC AUC values for UNSW-NB15}
\label{tab:UNSW}
\resizebox{\textwidth}{!}{%
\begin{tabular}{|l|l|c|c|c|c|c|c|c|c|c|c|c|c|c|c|c|c|}
\hline
\multirow{2}{*}{Detection approach} & \multirow{2}{*}{\begin{tabular}[c]{@{}l@{}}Modification\\ method\end{tabular}} & \multicolumn{2}{c|}{Fuzzers} & \multicolumn{2}{c|}{Analysis} & \multicolumn{2}{c|}{DoS} & \multicolumn{2}{c|}{Exploits} & \multicolumn{2}{c|}{Generic} & \multicolumn{2}{c|}{Reconnaissance} & \multicolumn{2}{c|}{Shellcode} & \multicolumn{2}{c|}{Worms} \\ \cline{3-18} 

 &  & \multicolumn{1}{c|}{PR} & \multicolumn{1}{c|}{ROC} & \multicolumn{1}{c|}{PR} & \multicolumn{1}{c|}{ROC} & \multicolumn{1}{c|}{PR} & \multicolumn{1}{c|}{ROC} & \multicolumn{1}{c|}{PR} & \multicolumn{1}{c|}{ROC} & \multicolumn{1}{c|}{PR} & \multicolumn{1}{c|}{ROC} & \multicolumn{1}{c|}{PR} & \multicolumn{1}{c|}{ROC} & \multicolumn{1}{c|}{PR} & \multicolumn{1}{c|}{ROC} & \multicolumn{1}{c|}{PR} & \multicolumn{1}{c|}{ROC} \\ \hline
 
Stand-alone LOF & None & 0.793 & 0.562 & 0.983 & 0.702 & \cellcolor[HTML]{C0C0C0}0.926 & \cellcolor[HTML]{C0C0C0}0.778 & 0.667 & 0.603 & 0.583 & 0.499 & 0.872 & 0.590 & 0.986 & 0.618 & \cellcolor[HTML]{C0C0C0}1.000 & 0.846 \\ \hline

AE-RE & None & 0.722 & 0.329 & 0.97 & 0.593 & 0.853 & 0.584 & 0.605 & 0.443 & 0.687 & 0.624 & 0.873 & 0.572 & 0.981 & 0.589 & 0.998 & 0.091 \\ \hline

\multirow{3}{*}{AE-LOF} & None & 0.786 & 0.520 & 0.986 & 0.720 & 0.862 & 0.589 & 0.633 & 0.506 & 0.563 & 0.463 & 0.869 & 0.557 & 0.987 & 0.627 & \cellcolor[HTML]{C0C0C0}1.000 & 0.877 \\ \cline{2-18} 
 & Pruning & 0.791 & 0.568 & 0.983 & 0.721 & 0.863 & 0.569 & 0.654 & 0.537 & 0.574 & 0.518 & 0.850 & 0.475 & 0.980 & 0.517 & \cellcolor[HTML]{C0C0C0}1.000 & 0.881 \\ \cline{2-18} 
 & Pruning+DA & 0.646 & 0.205 & 0.940 & 0.381 & 0.815 & 0.422 & 0.569 & 0.374 & 0.377 & 0.012 & 0.858 & 0.538 & \cellcolor[HTML]{C0C0C0}0.996 & \cellcolor[HTML]{C0C0C0}0.850 & \cellcolor[HTML]{C0C0C0}1.000 & 0.750 \\ \hline
 
\multirow{3}{*}{AEGR-LOF} & None & 0.785 & 0.532 & \cellcolor[HTML]{C0C0C0}0.998 & 0.948 & 0.858 & 0.601 & 0.620 & 0.499 & 0.597 & 0.518 & 0.858 & 0.538 & 0.988 & 0.632 & \cellcolor[HTML]{C0C0C0}1.000 & 0.861 \\ \cline{2-18} 
 & Pruning & 0.810 & 0.574 & \cellcolor[HTML]{C0C0C0}0.998 & \cellcolor[HTML]{C0C0C0}0.955 & 0.848 & 0.543 & 0.601 & 0.475 & 0.674 & 0.626 & 0.841 & 0.467 & 0.979 & 0.485 & \cellcolor[HTML]{C0C0C0}1.000 & \cellcolor[HTML]{C0C0C0}0.902 \\ \cline{2-18} 
 & Pruning+DA & \cellcolor[HTML]{C0C0C0}0.833 & \cellcolor[HTML]{C0C0C0}0.576 & 0.986 & 0.698 & 0.876 & 0.697 & \cellcolor[HTML]{C0C0C0}0.742 & \cellcolor[HTML]{C0C0C0}0.655 & \cellcolor[HTML]{C0C0C0}0.879 & \cellcolor[HTML]{C0C0C0}0.758 & \cellcolor[HTML]{C0C0C0}0.945 & \cellcolor[HTML]{C0C0C0}0.772 & 0.991 & 0.726 & 0.999 & 0.342\\ \hline
 
\end{tabular}%
}
\end{table*}
\begin{table*}
\centering
\caption{PR AUC and ROC AUC for NSL-KDD and CTU13}
\label{tab:NSLKDD}
\resizebox{\textwidth}{!}{%
\begin{tabular}{|l|l|c|c|c|c|c|c|c|c|c|c|c|c|c|c|c|c|}
\hline
\multirow{2}{*}{Detection approach} & \multirow{2}{*}{\begin{tabular}[c]{@{}l@{}}Modification\\ method\end{tabular}} & \multicolumn{2}{c|}{DoS} & \multicolumn{2}{c|}{Probe} & \multicolumn{2}{c|}{R2L} & \multicolumn{2}{c|}{U2R} & \multicolumn{2}{c|}{\begin{tabular}[c]{@{}c@{}}Virut\\ CTU13-13\end{tabular}} & \multicolumn{2}{c|}{\begin{tabular}[c]{@{}c@{}}Rbot\\ CTU13-10\end{tabular}} & \multicolumn{2}{c|}{\begin{tabular}[c]{@{}c@{}}Neris\\ CTU13-09\end{tabular}} & \multicolumn{2}{c|}{\begin{tabular}[c]{@{}c@{}}Murlo\\ CTU13-08\end{tabular}} \\ \cline{3-18} 
 &  & \multicolumn{1}{c|}{PR} & \multicolumn{1}{c|}{ROC} & \multicolumn{1}{c|}{PR} & \multicolumn{1}{c|}{ROC} & \multicolumn{1}{c|}{PR} & \multicolumn{1}{c|}{ROC} & \multicolumn{1}{c|}{PR} & \multicolumn{1}{c|}{ROC} & \multicolumn{1}{c|}{PR} & \multicolumn{1}{c|}{ROC} & \multicolumn{1}{c|}{PR} & \multicolumn{1}{c|}{ROC} & \multicolumn{1}{c|}{PR} & \multicolumn{1}{c|}{ROC} & \multicolumn{1}{c|}{PR} & \multicolumn{1}{c|}{ROC} \\ \hline
Stand-alone LOF & None & 0.549 & 0.555 & 0.830 & 0.635 & 0.810 & 0.687 & 0.994 & 0.530 & 0.644 & 0.781 & 0.845 & 0.465 & 0.708 & 0.060 & 0.940 & 0.594 \\ \hline

AE-RE & None & 0.547 & 0.440 & 0.800 & 0.056 & 0.790 & 0.543 & 0.994 & 0.569 & 0.431 & 0.463 & 0.815 & 0.057 & 0.864 & 0.003 & 0.920 & 0.001 \\ \hline

\multirow{3}{*}{AE-LOF} & None & 0.522 & 0.492 & 0.821 & 0.645 & 0.756 & 0.525 & 0.994 & 0.551 & 0.439 & 0.558 & 0.807 & 0.254 & 0.747 & 0.199 & 0.933 & 0.564 \\ \cline{2-18} 

 & Pruning & 0.549 & 0.512 & 0.895 & 0.742 & 0.757 & 0.409 & 0.994 & 0.516 & 0.371 & 0.412 & 0.799 & 0.241 & 0.981 & 0.880 & \cellcolor[HTML]{C0C0C0}0.988 & \cellcolor[HTML]{C0C0C0}0.908 \\ \cline{2-18} 
 
 & Pruning+DA & 0.687 & 0.781 & 0.683 & 0.296 & 0.766 & 0.548 & 0.993 & 0.451 & 0.518 & 0.639 & 0.895 & 0.523 & 0.966 & 0.900 & 0.800 & 0.068 \\ \hline
 
\multirow{3}{*}{AEGR-LOF} & None & 0.514 & 0.488 & 0.844 & 0.677 & 0.823 & 0.679 & 0.997 & 0.789 & 0.552 & 0.699 & 0.857 & 0.514 & 0.719 & 0.088 & 0.928 & 0.577 \\ \cline{2-18} 

 & Pruning & \cellcolor[HTML]{C0C0C0}0.925 & \cellcolor[HTML]{C0C0C0}0.883 & \cellcolor[HTML]{C0C0C0}0.947 & \cellcolor[HTML]{C0C0C0}0.841 & \cellcolor[HTML]{C0C0C0}0.859 & \cellcolor[HTML]{C0C0C0}0.685 & \cellcolor[HTML]{C0C0C0}0.998 & \cellcolor[HTML]{C0C0C0}0.817 & 0.472 & 0.588 & \cellcolor[HTML]{C0C0C0}0.899 & 0.547 & 0.891 & 0.452 & 0.979 & 0.782 \\ \cline{2-18} 
 
 & Pruning+DA & 0.643 & 0.695 & 0.797 & 0.598 & 0.674 & 0.287 & 0.993 & 0.550 & \cellcolor[HTML]{C0C0C0}0.797 & \cellcolor[HTML]{C0C0C0}0.832 & 0.859 & \cellcolor[HTML]{C0C0C0}0.597 & \cellcolor[HTML]{C0C0C0}0.992 & \cellcolor[HTML]{C0C0C0}0.985 & 0.917 & 0.332 \\ \hline
\end{tabular}%
}
\end{table*}

\begin{table*}
\centering
\caption{PR AUC and ROC AUC for data sets with single-type anomaly}
\label{tab:single-type}
\resizebox{\textwidth}{!}{%
\begin{tabular}{|l|l|c|c|c|c|c|c|c|c|c|c|}
\hline
\multirow{2}{*}{Detection approach} & \multirow{2}{*}{\begin{tabular}[c]{@{}l@{}}Modification\\ method\end{tabular}} & \multicolumn{2}{c|}{Spambase} & \multicolumn{2}{c|}{InternetAds} & \multicolumn{2}{c|}{PenDigits} & \multicolumn{2}{c|}{Shuttle} & \multicolumn{2}{c|}{Arrhythmia} \\ \cline{3-12} 
 &  & \multicolumn{1}{c|}{PR} & \multicolumn{1}{c|}{ROC} & \multicolumn{1}{c|}{PR} & \multicolumn{1}{c|}{ROC} & \multicolumn{1}{c|}{PR} & \multicolumn{1}{c|}{ROC} & \multicolumn{1}{c|}{PR} & \multicolumn{1}{c|}{ROC} & \multicolumn{1}{c|}{PR} & \multicolumn{1}{c|}{ROC} \\ \hline
 
Stand-alone LOF & None & 0.596 & 0.418 & 0.857 & 0.498 & 0.554 & 0.529 & 0.700 & 0.343 & 0.638 & 0.626 \\ \hline

AE-RE & None & 0.606 & 0.324 & 0.838 & 0.188 & 0.483 & 0.368 & 0.792 & 0.007 & 0.558 & 0.289 \\ \hline

\multirow{3}{*}{AE-LOF} & None & 0.579 & 0.428 & 0.853 & 0.515 & 0.574 & 0.590 & 0.819 & 0.689 & 0.681 & 0.678\\ \cline{2-12} 

 & Pruning & 0.650 & 0.497 & 0.909 & 0.638 & 0.980 & 0.970 & \cellcolor[HTML]{C0C0C0}0.839 & \cellcolor[HTML]{C0C0C0}0.720 & 0.703 & 0.688 \\ \cline{2-12} 
 
 & Pruning+DA & 0.503 & 0.334 & 0.859 & 0.538 & 0.315 & 0.078 & 0.635 & 0.105 & 0.605 & 0.616 \\ \hline
 
\multirow{3}{*}{AEGR-LOF} & None & 0.551 & 0.377 & 0.854 & 0.521 & 0.491 & 0.499 & 0.689 & 0.398 & 0.698 & 0.671 \\ \cline{2-12} 

 & Pruning & 0.558 & 0.388 & \cellcolor[HTML]{C0C0C0}0.928 & \cellcolor[HTML]{C0C0C0}0.687 & \cellcolor[HTML]{C0C0C0}0.998 & \cellcolor[HTML]{C0C0C0}0.998 & 0.696 & 0.414 & \cellcolor[HTML]{C0C0C0}0.728 & \cellcolor[HTML]{C0C0C0}0.732 \\ \cline{2-12} 
 
 & Pruning+DA & \cellcolor[HTML]{C0C0C0}0.682 & \cellcolor[HTML]{C0C0C0}0.646 & 0.854 & 0.538 & 0.970 & 0.958 & 0.812 & 0.559 & 0.512 & 0.484 \\ \hline
\end{tabular}%
}
\end{table*}
\par The performance of AEGR-LOF was compared with three different approaches. Besides employing the traditional LOF, an autoencoder with LOF (AE-LOF) and an autoencoder with RE (AE-RE) were used. In AE-LOF, the network is not using gradient reversal and its bottleneck is fed to the next stage for separating normal instances from anomalies. In AE-RE, the reconstruction error of the AE is used for detecting anomalies. It is worth mentioning that a denoising AE was not used in the experiment as it needs an anomaly-free training set and the purpose of this research is to stay needless of a clean training set. Also, it is possible to achieve higher performance by employing LOF if only a clean training set is used in advance to fit LOF first and use it as a one-class classifier \cite{Cao20193074}, which has already been done in the literature; however, in this work, LOF is trained based on the data which is extracted from the AE.

\par As mentioned earlier, the evaluation metrics used here are PR AUC and ROC AUC. The ROC AUC shows the area under the receiver operating characteristic curve (ROC) which presents the true positive rate versus the false negative rate at various thresholds. The range of AUC is $[0,1]$ where any value closer to $1$ shows that the model is performing better. In PR AUC, the true negatives have no impact on the metric. Instead, it reports the relationship between precision and recall at various thresholds. Similar to ROC AUC, the range is $[0,1]$ in which values closer to $1$ indicate a better performance. 

\par It can be seen in Table \ref{tab:UNSW} that the proposed model almost outperformed other approaches in every scenario (i.e., different types of network attack) in terms of both metrics. Stand alone LOF only showed superior results when applied to detect DoS attacks while also showed good results alongside other approaches when used to identify Worms. The UNSWB-NB15 dataset is a high dimensional dataset and the results can be used to support the assumption that LOF performs poorly when applied to this type of dataset. The NSL-KDD and CTU13 are two network datasets that are widely used in the literature. As shown in Table \ref{tab:NSLKDD}, the proposed model outperformed other approaches by a good margin except in one case, i.e., when applied to CTU13-08. 

\par Table \ref{tab:single-type} shows the performance when applied to datasets with a single type of anomaly. Based on the results, the proposed approach outperformed other models except when working on the Shuttle dataset. In particular, AEGR-LOF produced higher PR AUC and ROC AUC when detecting anomalies from the two single-type network datasets, i.e., Spambase and InternetAds. 

\begin{figure*}
\begin{subfigure}{.20\textwidth}
  \centering
  \includegraphics[width=1.2\linewidth]{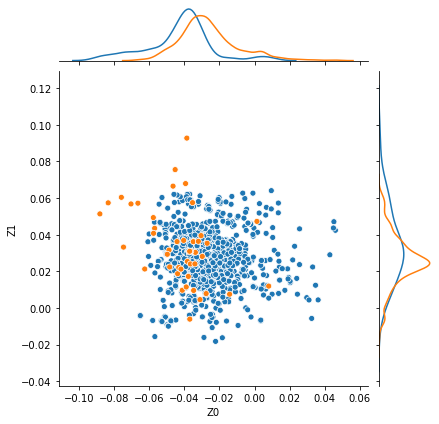}  
  \caption{AE}
  \label{fig:lv-first}
\end{subfigure}
\hfil
\begin{subfigure}{.20\textwidth}
  \centering
  \includegraphics[width=1.2\linewidth]{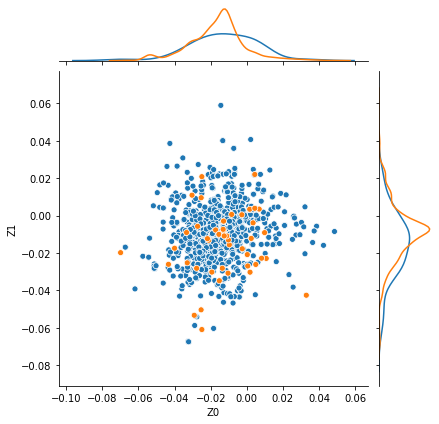}  
  \caption{AEGR}
  \label{fig:lv-second}
\end{subfigure}
\hfil
\begin{subfigure}{.20\textwidth}
  \centering
  \includegraphics[width=1.2\linewidth]{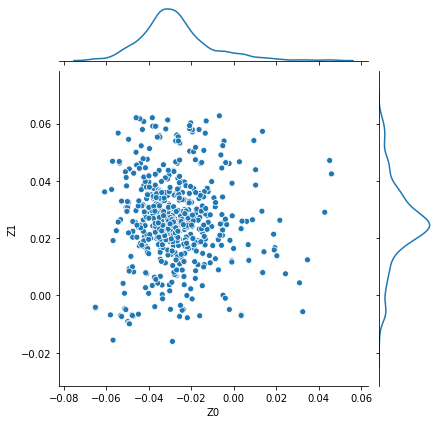}  
  \caption{AE (Pruned)}
  \label{fig:lv-third}
\end{subfigure}
\hfil
\begin{subfigure}{.20\textwidth}
  \centering
  \includegraphics[width=1.2\linewidth]{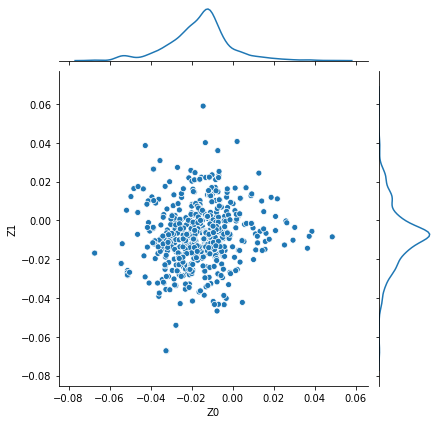}  
  \caption{AEGR (Pruned)}
  \label{fig:lv-fourth}
\end{subfigure}
\caption{Visualisation plot of the latent variables (the first two features) extracted by a simple AE and also AEGR on InternetAds. The orange points represent anomalies while the blue points represent the normal points. The curves at the top and right side of each figure indicates the KDE curves.}
\label{fig:latent_variables}
\end{figure*}

\par Figure \ref{fig:latent_variables} illustrates the latent variables produced by the simple AE and the AEGR model. While Figure \ref{fig:lv-first} and Figure \ref{fig:lv-third} show the latent variables extracted from the bottleneck of the simple AE, Figure \ref{fig:lv-second} and Figure \ref{fig:lv-fourth} present the latent variables generated by the proposed model. By looking at Figure \ref{fig:lv-first} and Figure \ref{fig:lv-second}, which show the latent variables before pruning, it can be seen that while the simple AE managed to regenerate some anomalies correctly and separate them from the normal data points, the AEGR model failed at separating anomalies due to the constraint applied, i.e., they stayed close to the dense area. However, this failure means that the AEGR model should have generated high reconstruction errors for anomalies. After applying a simple pruning based on the reconstruction error (Figure \ref{fig:lv-third} and Figure \ref{fig:lv-fourth}), the latent variables created by the AEGR model made a denser cluster compared to the simple AE, which led to getting a better performance from LOF as it can define a better class boundary when the training set is very dense and homogeneous.

\par Overall, AEGR-LOF with pruning or pruning and data augmentation achieved 17 best results from 22 comparisons. Therefore, it can be concluded that AEGR-LOF is significantly better than other approaches, which shows the importance of regularisation in AEs when used in anomaly detection problems. To support this conclusion further, Wilcoxon signed-rank test \cite{kerby2014simple} was carried out to verify whether the improvement was significant or not. Therefore, the NSL-KDD(DoS) dataset was randomly selected and the test was carried out on both LOF and AEGR-LOF with pruning and data augmentation $20$ times, i.e., $N=20$. By feeding PR AUCs to the Wilcoxon test, it was confirmed that the results are significantly different. It is worth recalling that carrying out repeated Wilcoxon tests on multiple algorithms is not recommended as it increases the chance of rejecting a certain proportion of the null hypotheses merely based on random chance \cite{demvsar2006statistical}.


\section{Conclusion}
\label{sec:conclusion}
\par In this paper, a new model is proposed to detect network anomalies, particularly in large datasets, that traditional algorithms such as LOF are incapable of dealing with. In the proposed approach, a novel autoencoder called AEGR is utilised to reduce the dimensionality of large datasets, transforming the data into a lower dimensional space while minimising the loss of vital features. Normal AEs fail to produce satisfactory results when the data is polluted with noise and anomalies because the network learns to replicate them together with normal data instances. Unlike other approaches that either try to use an insensitive loss function or train the network by injecting noise, the unsupervised model presented in this work, at each epoch, finds the data instances that caused the highest weight update, and then manipulates the inverted backpropagated gradients to counter that update. Finally, we apply the original LOF to the extracted features from the bottleneck of the autoencoder in order to separate normal instances from anomalies. Based on the results that were achieved from the experiments conducted on seven datasets, it was shown that the AEGR-LOF model is capable of achieving better results compared to the traditional LOF and other similar approaches such as a simple Autoencoder followed by a threshold-based classifier. The performance of the proposed model was evaluated using two metrics in which overall the AEGR model showed superior results.

\bibliographystyle{ieeetr}
\bibliography{references.bib}
\end{document}